\documentclass[10pt,twocolumn,letterpaper]{article}

\usepackage{cvpr}
\usepackage{times}
\usepackage{epsfig}
\usepackage{graphicx}
\usepackage{float}
\usepackage{amsmath}
\usepackage{amssymb}
\usepackage{verbatim}
\usepackage{subcaption} 
\usepackage{booktabs}
\usepackage{cuted}
\usepackage{capt-of}
\usepackage{url}
\usepackage{arydshln}
\usepackage{bm}

\usepackage[pagebackref=true,breaklinks=true,letterpaper=true,colorlinks,bookmarks=false]{hyperref}

\cvprfinalcopy % *** Uncomment this line for the final submission

\def\cvprPaperID{1691} % *** Enter the CVPR Paper ID here

\ifcvprfinal\fi
\begin{document}

%%%%%%%%% TITLE
%\title{Automatic Music Video Generation with pose perceptual loss}
\title{Music-oriented Dance Video Synthesis with Pose Perceptual Loss}

\author{Xuanchi Ren
\qquad
Haoran Li
\qquad
Zijian Huang
\qquad
Qifeng Chen
\vspace{1em}
\\HKUST
}

\twocolumn[{%
\renewcommand\twocolumn[1][]{#1}%
\maketitle
\begin{center}
    \centering
    \begin{tabular}{@{}c@{\hspace{0.1em}}c@{\hspace{0.1em}}c@{\hspace{0.1em}}c@{\hspace{0.1em}}c@{}}
        
         \includegraphics[width=0.19\textwidth]{{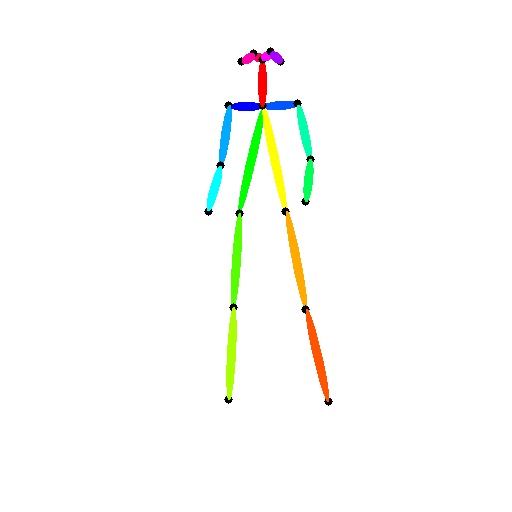}} &
         \includegraphics[width=0.19\textwidth]{{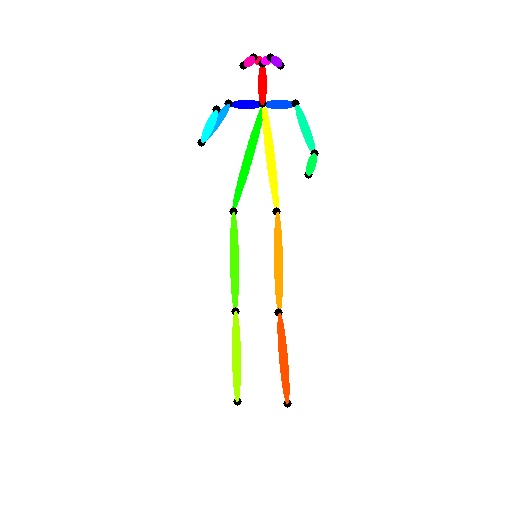}} &
         \includegraphics[width=0.19\textwidth]{{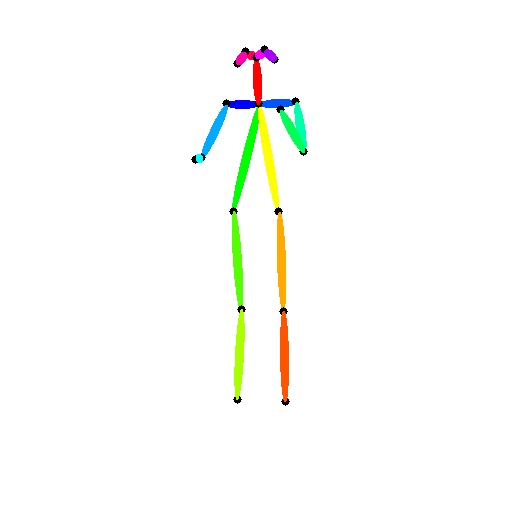}} &
         \includegraphics[width=0.19\textwidth]{{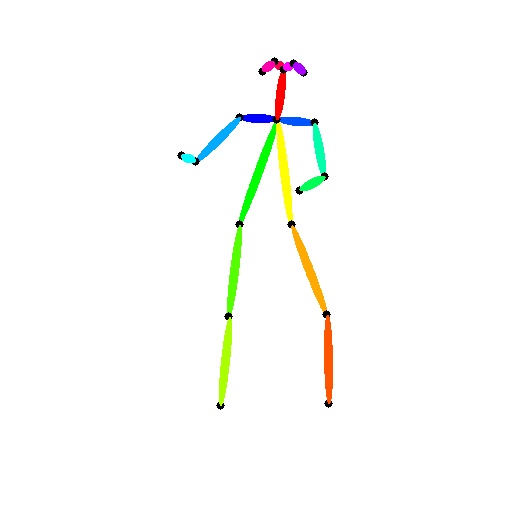}} &
         \includegraphics[width=0.19\textwidth]{{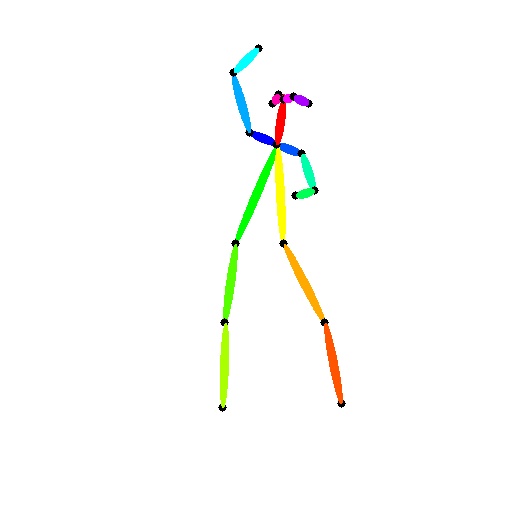}} \\
         \includegraphics[width=0.19\textwidth]{{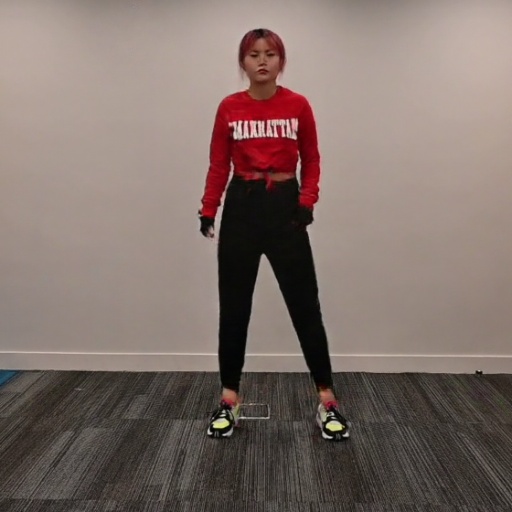}} &
         \includegraphics[width=0.19\textwidth]{{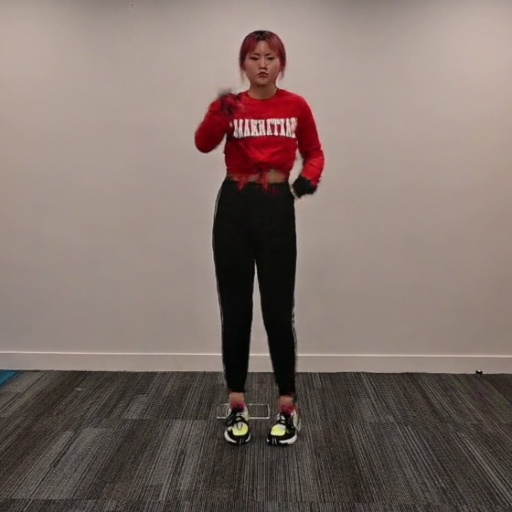}} &
         \includegraphics[width=0.19\textwidth]{{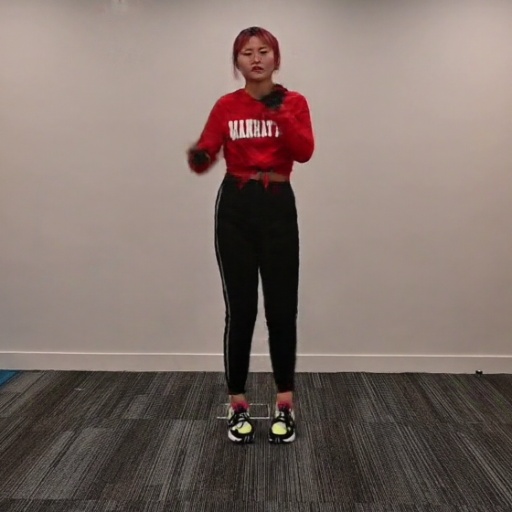}} &
         \includegraphics[width=0.19\textwidth]{{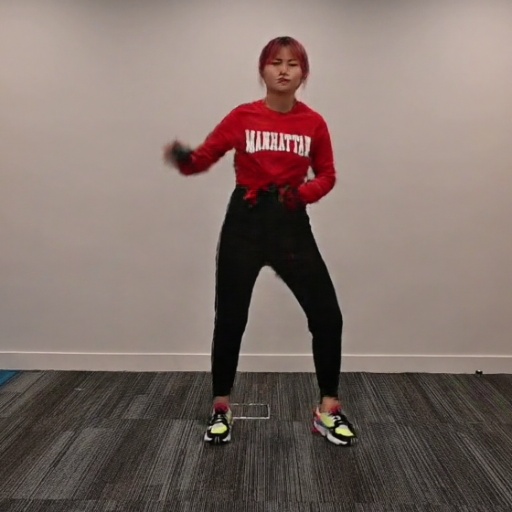}} &
         \includegraphics[width=0.19\textwidth]{{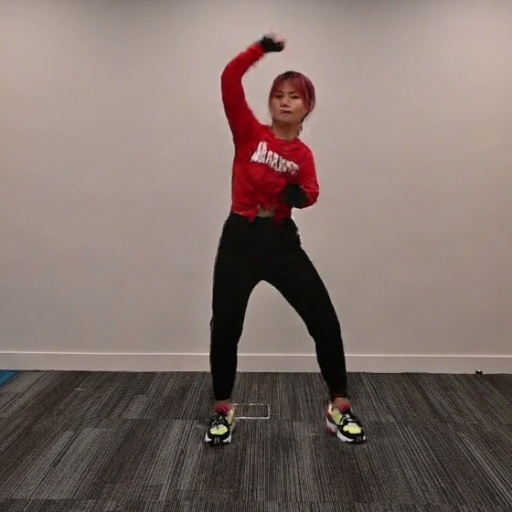}} \\
        %   \label{fig:dn_Demo}
     \end{tabular} 
  \captionof{figure}{Our synthesized dance video conditioned on the music \textsl{``I Wish"}. We show 5 frames from a 5-second synthesized video. The top row shows the skeletons, and the bottom row shows the corresponding synthesized video frames. More results are shown in the supplementary video at \url{https://youtu.be/0rMuFMZa_K4}.}
  \label{fig:dn_Demo}
\end{center}
}]

%\thispagestyle{empty}

%%%%%%%%% ABSTRACT
\begin{abstract}
    We present a learning-based approach with pose perceptual loss for automatic music video generation. Our method can produce a realistic dance video that conforms to the beats and rhymes of almost any given music. To achieve this, we firstly generate a human skeleton sequence from music and then apply the learned pose-to-appearance mapping to generate the final video. In the stage of generating skeleton sequences, we utilize two discriminators to capture different aspects of the sequence and propose a novel pose perceptual loss to produce natural dances. Besides, we also provide a new cross-modal evaluation to evaluate the dance quality, which is able to estimate the similarity between two modalities of music and dance. Finally, a user study is conducted to demonstrate that dance video synthesized by the presented approach produces surprisingly realistic results. Source code and data are available at \url{https://github.com/xrenaa/Music-Dance-Video-Synthesis}.
\end{abstract}

%%%%%%%%% BODY TEXT
\vspace{-1em}
\section{Introduction}
Music videos have become unprecedentedly popular all over the world. Nearly all the top 10 most-viewed YouTube videos\footnote{\url{https://www.digitaltrends.com/web/most-viewed-youtube-videos/}} are music videos with dancing. While these music videos are made by professional artists, we wonder if an intelligent system can automatically generate personalized and creative music videos. In this work, we study automatic dance music video generation, given almost any music. We aim to synthesize a coherent and photo-realistic dance video that conforms to the given music. With such music video generation technology, a user can share a personalized music video on social media. In Figure \ref{fig:dn_Demo}, we show some images of our synthesized dance video given the music \textsl{``I Wish"} by \textsl{Cosmic Girls}.

The dance video synthesis task is challenging for various technical reasons. Firstly, the mapping between dance motion and background music is ambiguous: different artists may compose distinctive dance motion given the same music. This suggests that a simple machine learning model with $L_1$ or $L_2$ distance~\cite{Leelisten,tang2018dance} can hardly capture the relationship between dance and music. Secondly, it is technically difficult to model the space of human body dance. The model should avoid generating non-natural dancing movements. Even slight deviations from normal human poses could appear unnatural. Thirdly, no high-quality dataset is available for our task. Previous motion datasets~\cite{motion_dataset1,motion_dataset2} mostly focus on action recognition. Tang et al.~\cite{tang2018dance} provide a 3D joint dataset for our task. However, we encounter errors that the dance motion and the music are not aligned when we try to use it.

Nowadays, there are a large number of music videos with dancing online, which can be used for the music video generation task. To build a dataset for our task, we apply OpenPose~\cite{cao2018openpose,cao2017realtime,wei2016cpm} to get dance skeleton sequences from online videos. However, the skeleton sequences acquired by OpenPose are very noisy: some estimated human poses are inaccurate. Correcting such a dataset is time-consuming by removing inaccurate poses and thus not suitable for extensive applications. Furthermore, only $L_1$ or $L_2$ distance is used for training a network in prior work~\cite{Leelisten,tang2018dance,Yalta}, which is demonstrated to disregard some specific motion characteristics by ~\cite{MartinezB017}. To tackle these challenges, we propose a novel pose perceptual loss so that our model can be trained on noisy data (imperfect human poses) gained by OpenPose.

Dance synthesis has been well studied in the literature by searching dance motion in a database using music as a query~\cite{alemi2017groovenet, Kim2006MakingTD, Shiratori2006DancingtoMusicCA}. These approaches can not generalize well to music beyond the training data and lack creativity, which is the most indispensable factor of dance. To overcome such obstacles, we choose the generative adversarial network (GAN)~\cite{goodfellow2014generative} to deal with cross-modal mapping. However, Cai et al.~\cite{cai2018deep} showed that human pose constraints are too complicated to be captured by an end-to-end model trained with a direct GAN method. Thus, we propose to use two discriminators that focus on local coherence and global harmony, respectively. 

In summary, the contributions of our work are:
\begin{itemize}
\setlength{\itemsep}{0pt}
\setlength{\parsep}{0pt}
\setlength{\parskip}{0pt}
\item With the proposed pose perceptual loss, our model can be trained on a noisy dataset (without human labels) to synthesize realistic dance video that conforms to almost any given music.
\item With the Local Temporal Discriminator and the Global Content Discriminator, our framework can generate a coherent dance skeleton sequence that matches the length, rhythm, and the emotion of music.
\item For our task, we build a dataset containing paired music and skeleton sequences, which will be made public for research. To evaluate our model,  we also propose a novel cross-modal evaluation that measures the similarity between music and a dance skeleton sequence.
\end{itemize}

\section{Related Work}
% \textbf{Pose Recognition} 
% ~\cite{yan2018spatial}
% ~\cite{li2018co}

\begin{figure*}[t]
\centering
   \includegraphics[width=\linewidth]{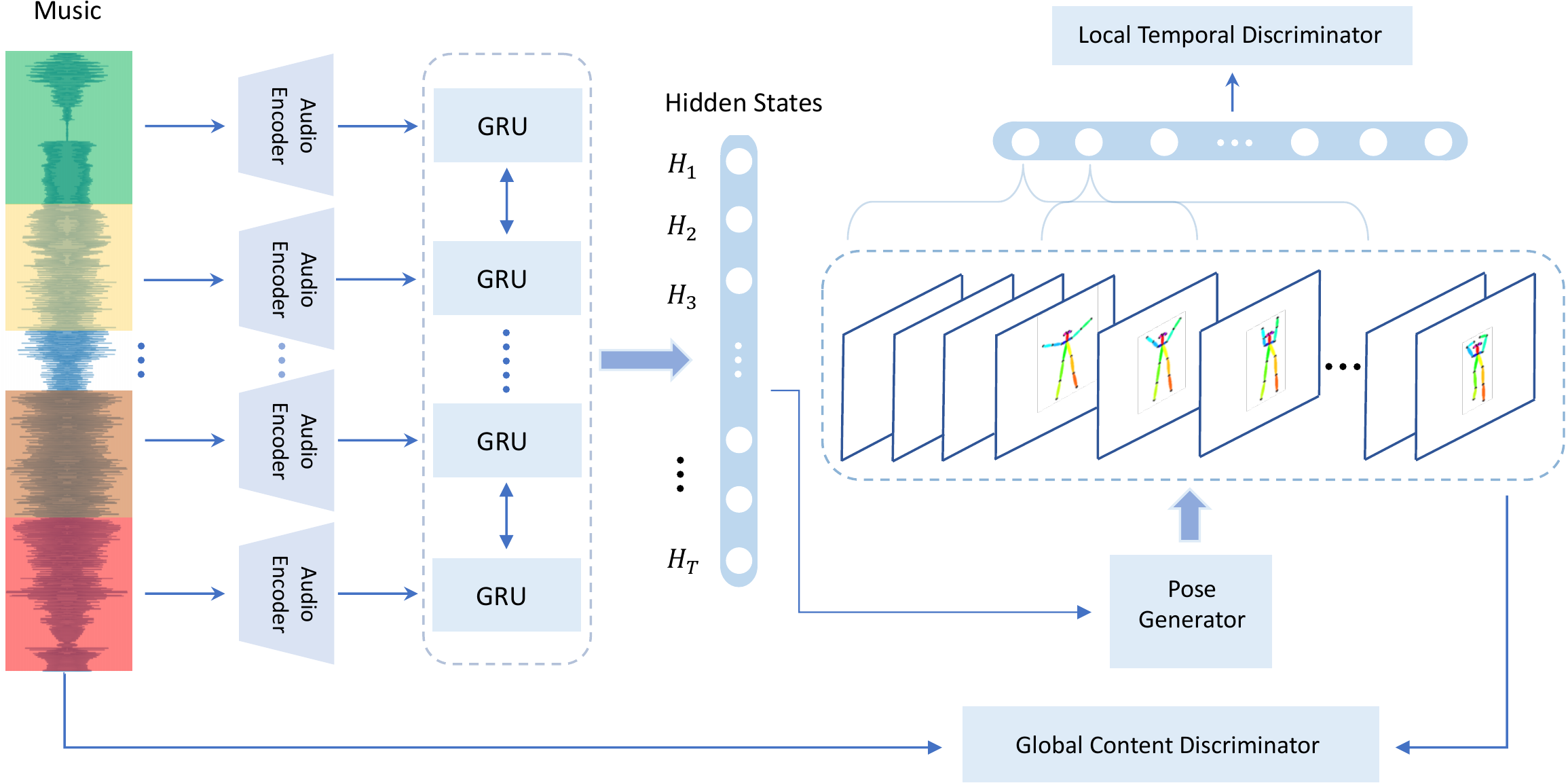}
   \vspace{0.5em}
   \caption{Our framework for human skeleton sequence synthesis. The input is music signals, which are divided into pieces of 0.1-second music. The generator contains an audio encoder, a bidirectional GRU, and a pose generator. The output skeleton sequence of the generator is fed into the Global Content Discriminator with the music. The generated skeleton sequence is then divided into overlapping sub-sequences, which are fed into the Local Temporal Discriminator.}
\label{fig:Overview}
\vspace{0.5em}
\end{figure*}

\textbf{GAN-based Video Synthesis.} 
A generative adversarial network (GAN)~\cite{goodfellow2014generative} is a popular approach for image generation. The images generated by GAN are usually sharper and with more details compared to those with $L_1$ and $L_2$ distance. Recently, GAN is also extended to video generation tasks~\cite{Li2017VideoGF,Mathieu2015DeepMV,Tulyakov:2018:MoCoGAN,Vondrick}. The most simple changes made in GANs for videos are proposed in ~\cite{TGAN2017, Vondrick}. The GAN model in ~\cite{Vondrick} replaced the standard 2D convolutional layer with a 3D convolutional layer to capture the temporal feature, although this characteristic capture method is limited in the fixed time. TGAN~\cite{TGAN2017} overcame the limitation but with the cost of constraints imposed in the latent space. MoCoGAN~\cite{Tulyakov:2018:MoCoGAN} could generate videos that combine the advantages of RNN-based GAN models and sliding window techniques so that the motion and content are disentangled in the latent space.

Another advantage of GAN models is that it is widely applicable to many tasks, including the cross-modal audio-to-video problem. Chen et al.~\cite{Audio-Visual-Integration} proposed a GAN-based encoder-decoder architecture using CNNs to convert between audio spectrograms and frames. Furthermore, Vougioukas et al.~\cite{vougioukas2018end} adapted temporal GAN to synthesize a talking character conditioned on speech signals automatically.

\textbf{Dance Motion Synthesis.} 
A line of work focuses on the mapping between acoustic and motion features. On the base of labeling music with joint positions and angles, Shiratori et al.~\cite{Shiratori2006DancingtoMusicCA,Kim2006MakingTD} incorporated gravity and beats as additional features for predicting dance motion. Recently, Alemi et al.~\cite{alemi2017groovenet} proposed to combine the acoustic feature with the motion features of previous frames. However, these approaches are entirely dependent on the prepared database and may only create rigid motion when it comes to music with similar acoustic features.

Recently, Yaota et al.~\cite{Yalta} accomplished dance synthesis using standard deep learning models. The most recent work is by Tang et al.~\cite{tang2018dance}, who proposed a model based on LSTM-autoencoder architecture to generate dance pose sequences. Their approach is trained with a  $L_2$ distance loss, and their evaluation only includes comparisons with randomly sampled dances that are not on a par with those by real artists. Their approach may not work well on the noisy data obtained by OpenPose.   

\section{Overview}
To generate a dance video from music, we split our system into two stages. In the first stage, we propose an end-to-end model that directly generates a dance skeleton sequence according to the audio input. In the second stage, we apply an improved pix2pixHD GAN~\cite{Wang0ZTKC18,chan2018everybody} to transfer the dance skeleton sequence to a dance video. In this overview, we will mainly describe the first stage, as shown in Figure \ref{fig:Overview}.

Let $V$ be the number of joints of the human skeleton, and the dimension of a 2D coordinate $(x,y)$ is
2. We formulate a dance skeleton sequence $X$ as a sequence of human skeletons across $T$ consecutive frames in total: $X \in R^{ T\times 2V }$ where each skeleton frame $X_{t}\in R^{2V}$ is a vector containing all $(x,y)$ joint locations. Our goal is to learn a function $G: R^{TS} \rightarrow R^{T\times 2V}$ that maps audio signals with sample rate $S$ per frame to a joint location vector sequence.

\textbf{Generator.}
The generator is composed of a music encoding part and a pose generator. The input audio signals are divided into pieces of 0.1-second music. These pieces are encoded using 1D convolution and then fed into a bi-directional 2-layer GRU in chronological order, resulting in output hidden states $O = \{H_{1},H_{2}, \cdots ,H_{T} \}$. These hidden states are fed in the pose generator, which is a multi-layer perceptron to produce a skeleton sequence $X$.

\textbf{Local Temporal Discriminator.}
The output skeleton sequence $X$ is divided into $K$ overlapping sequences $\in R^{t\times 2V}$. Then these sub-sequences are fed into the Local Temporal Discriminator, which is a two-branch convolutional network. In the end, a small classifier outputs $K$ scores that determine the realism of these skeleton sub-sequences. 

\begin{figure*}[t]
\centering
   \includegraphics[width=\linewidth]{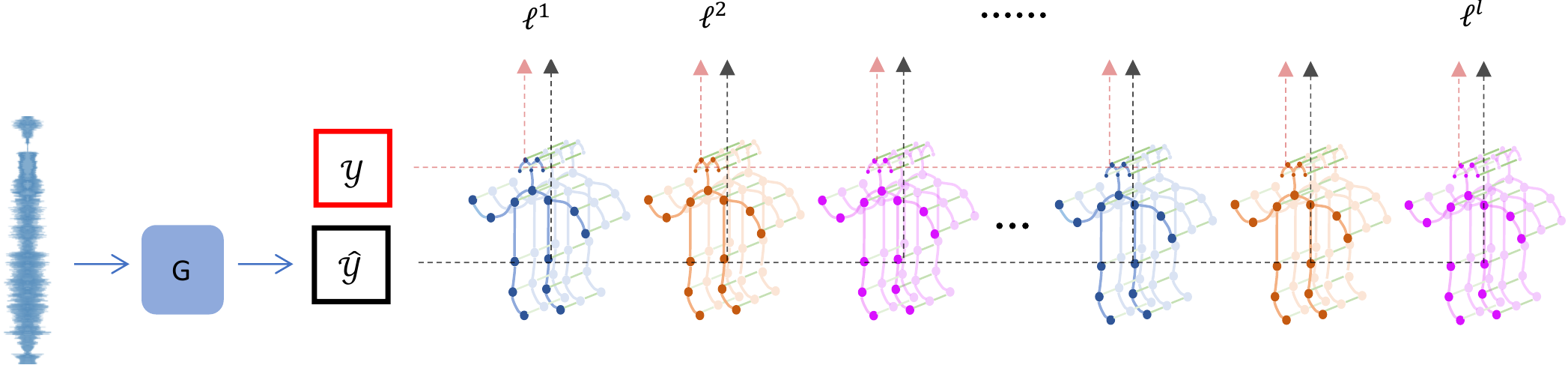}
   \caption{The overview of the pose perceptual loss based on ST-GCN. $G$ is our generator in the first stage. $y$ is the ground-truth skeleton sequence, and $\hat{y}$ is the generated skeleton sequence.}
\label{fig:GCN}
\end{figure*}

\textbf{Global Content Discriminator.}
The input to the Global Content Discriminator includes the music $M\in R^{TS}$ and the dance skeleton sequence $X$. For the pose part, the skeleton sequence $X$ is encoded using pose discriminator as $F^{P}\in R^{256}$. For the music part, similar to the sub-network of the generator, music is encoded using 1D convolution and then fed into a bi-directional 2-layer GRU, resulting an output $O^{M} = \{H^{M}_{1}, H^{M}_{2},..., H^{M}_{T} \}$ and $O^{M}$ is transmitted into the self-attention component of ~\cite{lin2017structured} to get a comprehensive music feature expression $F^{M}\in R^{256}$. In the end, we concatenate $F^M$ and $F^P$ along channels and use a small classifier, composed of a 1D convolutional layer and a fully-connected (FC) layer, to determine if the skeleton sequence matches the music.

\textbf{Pose Perceptual Loss.}
Recently,  Graph Convolutional Network (GCN) has been extended to model skeletons since the human skeleton structure is graph-structured data. Thus, the feature extracted by GCN remains a high-level spatial structural information between different body parts. Matching activations in a pre-trained GCN network gives a better constraint on both the detail and layout of a pose than the traditional methods such as $L_1$ distance and $L_2$ distance. Figure \ref{fig:GCN} shows the pipeline of the pose perceptual loss. With the pose perceptual loss, our output skeleton sequences do not need an additional smooth step or any other post-processing.

\section{Pose Perceptual Loss}
Perceptual loss or feature matching loss ~\cite{BrunaSL15,ChenK17,DosovitskiyB16,JohnsonAF16,NguyenDYBC16,Wang0ZTKC18,wang2018video} is a popular loss to measure the similarity between two images in image processing and synthesis. For the tasks that generate human skeleton sequences~\cite{cai2018deep,Leelisten,tang2018dance}, only $L_1$ or $L_2$ distance is used for measuring pose similarity. With a $L_1$ or $L_2$ loss, we find that our model tends to generate poses conservatively (repeatedly) and fail to capture the semantic relationship across motion appropriately. Moreover, the datasets generated by OpenPose~\cite{cao2018openpose,cao2017realtime,wei2016cpm} are very noisy, as shown in Figure \ref{fig:K-pop}. Correcting inaccurate human poses on a large number of videos is labor-intensive and undesirable: a two-minute video with 10 FPS will have 1200 poses to verify. To tackle these difficulties, we propose a novel pose perceptual loss.

\begin{figure}[t]
\setlength{\tabcolsep}{0pt}
\centering
    \begin{tabular}{@{}c@{}c:c@{}c@{}}
         {\includegraphics[width=0.24\columnwidth]{{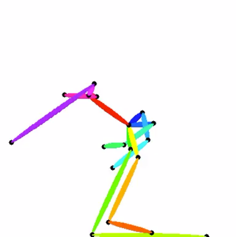}}} &
         {\includegraphics[width=0.24\columnwidth]{{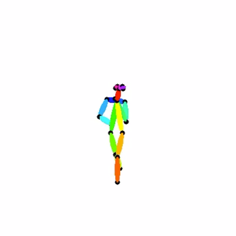}}} &
         {\includegraphics[width=0.24\columnwidth]{{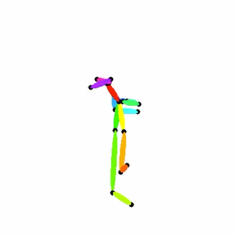}}} &
         {\includegraphics[width=0.24\columnwidth]{{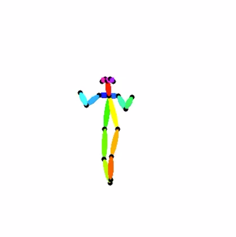}}} \\
     \end{tabular} 
     \caption{ In each section, the first image is a skeleton generated by the model without pose perceptual loss, and the second image is a skeleton generated by the model with pose perceptual loss according to the same piece of music. }
    \label{fig:unclean}
\end{figure}

The idea of perceptual loss is originally studied in the image domain, which is used to match activations in a visual perception network such as VGG-19~\cite{SimonyanZ14a,ChenK17}. To use the traditional perceptual loss, we need to draw generated skeletons on images, which is complicated and seemingly suboptimal. Instead of projecting pose joint coordinates to an image, we propose to directly match activations in a pose recognition network that takes human skeleton sequences as input. Such a network is mainly aimed at pose recognition or prediction tasks, and ST-GCN~\cite{yan2018spatial} is a Graph Convolutional Network (GCN) that is applicable to be a visual perception network in our case. ST-GCN utilizes a spatial-temporal graph to form the hierarchical representation of skeleton sequences and is capable of automatically learning both spatial and temporal patterns from data. To test the impact of the pose perceptual loss on our noisy dataset, we prepare a 20-video dataset with many noises due to the wrong pose detection of OpenPose. As shown in Figure \ref{fig:unclean}, our generator can stably generate poses with the pose perceptual loss.

Given a pre-trained GCN network $\Phi$, we define a collection of layers $\Phi$ as $\{ \Phi_{l} \}$. For a training pair $\left( P, M\right) $, where $P$ is the ground truth skeleton sequence and $M$ is the corresponding piece of music, our perceptual loss is
\begin{equation}
    \mathcal{L}_{P} = \sum_{l}\lambda_{l}\Arrowvert\Phi_{l}(P)-\Phi_{l}(G(M)) \Arrowvert_{1}.
\end{equation}
Here $G$ is the first-stage generator in our framework. The hyperparameters $\{\lambda_{l}\}$ balance the contribution of each layer $l$ to the loss.

\section{Implementation}
\subsection{Pose Discriminator}
To evaluate if a skeleton sequence is an excellent dance, we believe the most indispensable factors are the intra-frame representation for joint co-occurrences and the inter-frame representation for skeleton temporal evolution. To extract features of a pose sequence, we explore multi-stream CNN-based methods and adopt the Hierarchical Co-occurrence Network framework~\cite{li2018co} to enable discriminators to differentiable real and fake pose sequences.

\textbf{Two-Stream CNN.}
The input of the pose discriminator is a skeleton sequence $X$. The temporal difference is interpolated to be of the same shape of $X$. Then the skeleton sequence and the temporal difference are fed into the network directly as two streams of inputs. Their feature maps are fused by concatenation along channels, and then we use convolutional and fully-connected layers to extract features.

\subsection{Local Temporal Discriminator}
One of the objectives of the pose generator is the temporal coherence of the generated skeleton sequence. For example, when a man moves his left foot, his right foot should keep still for multiples frames.
Similar to PatchGAN~\cite{isola2017image,zhu2017unpaired,wang2018video}, we propose to use Local Temporal Discriminator, which is a 1D version of PatchGAN to achieve coherence between consecutive frames. Besides, the Local Temporal Discriminator contains a trimmed pose discriminator and a small classifier.
% To reach coherence and smoothness  of the sequence, Cai et al.~\cite{cai2018deep} conducted research on generating skeleton sequences from random noise by generating shifts between two consecutive frames. For our task, each frame corresponds to a piece of music piece, and thus it is not wise to generate the shift based on music.
% For image generation, PatchGAN~\cite{isola2017image,zhu2017unpaired,wang2018video} is commonly used to pay attention to the details of high-resolution images. 

\subsection{Global Content Discriminator}
Dance is closely related to music, and the harmony between music and dance is a crucial criterion to evaluate a dance sequence. Inspired by ~\cite{vougioukas2018end}, we proposed the Global Content Discriminator to deal with the relationship between music and dance. 

As we mentioned previously, music is encoded as a sequence $O^{M} = \{H^{M}_{1},H^{M}_{2},...,H^{M}_{T} \}$. 
% In the beginning, we use the last hidden state $H^{M}_{T}$ to represent music feature $F^{M}$ and concatenate it with pose feature  $F^{P}$ and then feed the combined feature to a classifier to determine if the skeleton sequence matches the music. 
Though GRU can capture long term dependencies, it is still challenging for GRU to encode the entire music information. In our experiment, only using $H^{M}_{T}$ to represent music feature $F^{M}$ will lead to a crash of the beginning part of the skeleton sequence. Therefore, we use the self-attention mechanism~\cite{lin2017structured} to assign a weight for each hidden state and gain a comprehensive embedding. In the next part, we briefly describe the self-attention mechanism used in our framework.

\textbf{Self-attention mechanism.}
 Given $O^{M    }\in R^{T\times k}$, we can compute its weight at each time step by
\begin{equation}
    r=W_{s2}\tanh(W_{s1}{O^{M}}^{\top}),
\end{equation}
\begin{equation}
     a_{i} = -\log \left( \frac{\exp \left(r_{i}\right)}{\sum_{j}\exp \left(r_{j}\right)}\right),
\end{equation}
where $r_{i}$ is $i$-th element of the $r$ while $W_{s1}\in R^{k\times l}$  and $W_{s2}\in R^{l\times 1}$. $a_i$ is the assigned weight for $i$-th time step in the sequence of hidden states. Thus, the music feature $F^{M}$ can be computed by multiplying the scores $A=[a_{1},a_{2},...,a_{n}]$ and $O^{M}$, written as $F^{M}=AO^{M}$.

\subsection{Other Loss Function}
\textbf{GAN loss}
$\mathcal{L}_{adv}\textbf{.}$
The Local Temporal Discriminator ($D_{local}$) is trained on overlapping skeleton sequences that are sampled using $S(\cdot)$ from a whole skeleton sequence. The Global Content Discriminator ($D_{global}$) distinguishes the harmony between the skeleton sequence and the input music $m$. Besides, we have $x = G(m)$ and the ground truth skeleton sequence $p$. We also apply a gradient penalty~\cite{wgan} term in $D_{global}$. Therefore, the adversarial loss is defined as
\begin{equation}
\begin{aligned}
    \mathcal{L}_{adv}=&\mathbb{E}_{p}[\log D_{local}(S(p))]+\\
    &\mathbb{E}_{x,m}[\log[1-D_{local}(S(x))]]+\\
    &\mathbb{E}_{p,m}[\log D_{global}(p,m)]+\\
    &\mathbb{E}_{x,m}[\log[1-D_{global}(x,m)]]+\\
    &w_{GP}\mathbb{E}_{x\hat{,}m}[(\Arrowvert \bigtriangledown_{x\hat{,}m}D(x\hat{,} m) \Arrowvert_{2}-1)^{2}].
\end{aligned}
\end{equation}
where $w_{GP}$ is the weight for the gradient penalty term.

\textbf{$L_1$ distance} $\mathcal{L}_{L_1}\textbf{.}$ 
Given a ground truth dance skeleton sequence $Y$ with the same shape of $X \in R^{ T\times 2V }$, the reconstruction loss at the joint level is:
\begin{equation}
    \mathcal{L}_{L_1}=\sum_{j\in [0,2V]} {\Arrowvert Y_j-X_j \Arrowvert}_{1}.
\end{equation}

\begin{figure}[t]
\centering
    \begin{tabular}{@{}c@{\hspace{0.1em}}c@{\hspace{0.1em}}c@{\hspace{0.1em}}c@{}}
         {\includegraphics[width=0.24\columnwidth]{{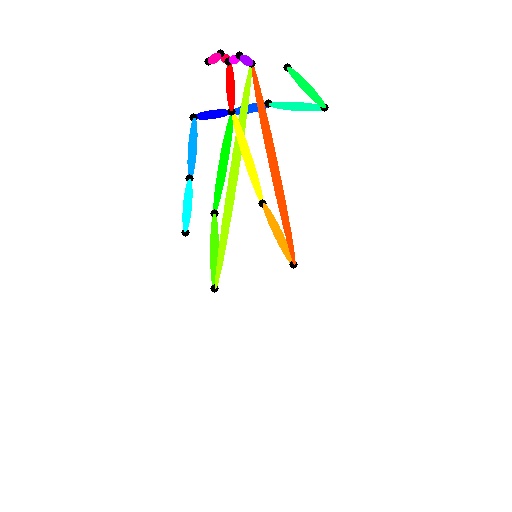}}} &
         {\includegraphics[width=0.24\columnwidth]{{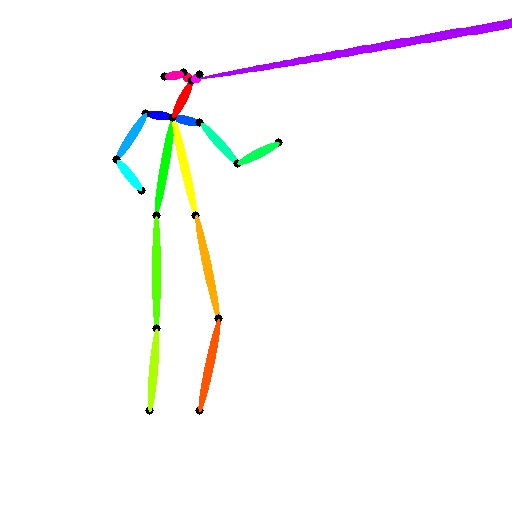}}} &
         {\includegraphics[width=0.24\columnwidth]{{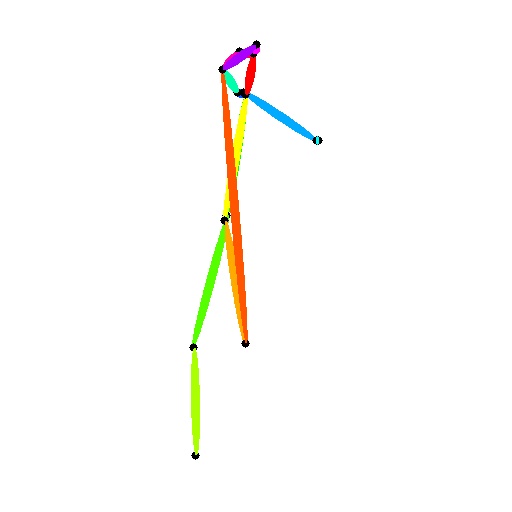}}} &
         {\includegraphics[width=0.24\columnwidth]{{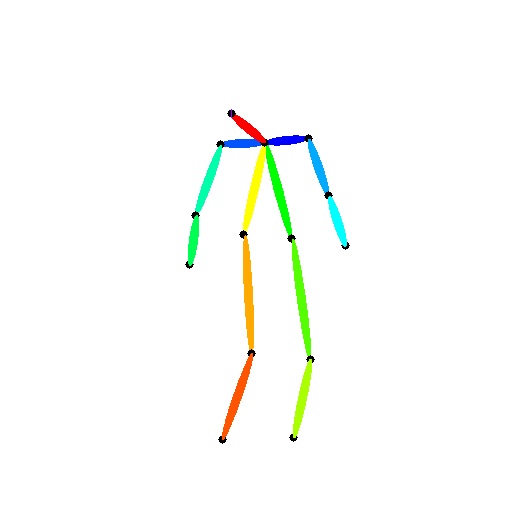}}} \\
     \end{tabular} 
     \caption{Noisy data caused by occlusion and overlapping. For the first part of the K-pop dataset, there is a large number of such skeletons. For the second part of the K-pop dataset, there are few inaccurate skeletons.}
    \label{fig:K-pop}
\end{figure}

\textbf{Feature matching loss}
$\mathcal{L}_{FM}\textbf{.}$ We adopt the feature matching loss from ~\cite{Wang0ZTKC18} to stabilize the training of Global Content Discriminator $D$:
 \begin{equation}
      \mathcal{L}_{FM}=\mathbb{E}_{p,m}\sum_{i=1}^{M} {\Arrowvert D^i(p,m)-D^i(G(m),m)\Arrowvert}_1.
 \end{equation}
where $M$ is the number of layers in $D$ and $D^i$ denotes the $i^{th}$ layer of $D$. In addition, we omit the normalization term of the original $\mathcal{L}_{FM}$ to fit our architecture.   

\textbf{Full Objective.}
Our full objective is
\begin{equation}
    \arg \min_{G} \max_{D}\mathcal{L}_{adv}+w_{P}\mathcal{L}_{P}+w_{FM}\mathcal{L}_{FM}+
    w_{L_1}\mathcal{L}_{L_1}.
\end{equation}
where $w_P$, $w_{FM}$, and $w_{L_1}$ represent the weights for each loss term.

\subsection{Pose to Video} 
Recently,  researchers have been studying motion transfer, especially for transferring dance motion between two videos~\cite{chan2018everybody, LiquidGAN,wang2018video,dancedance}. Among these methods, we adopt the approach proposed by Chan et al.~\cite{chan2018everybody} for its simplicity and effectiveness. Given a skeleton sequence and a video of a target person, the framework could transfer the movement of the skeleton sequence to the target person.
%by utilizing an improved pix2pixHD GAN～\cite{Wang0ZTKC18}. 
% They propose to applying OpenPose~\cite{cao2018openpose,cao2017realtime,wei2016cpm} to the source video and gain a skeleton sequence and then generate the dance video condition on the skeleton sequence without any other input such like Dense Human Pose~\cite{GulerNK18} and Optical Flow. In our task, by feeding a skeleton sequence and a target video to the pipeline in ~\cite{chan2018everybody},  we can get a dance video. 
We used a third-party implementation\footnote{\url{https://github.com/CUHKSZ-TQL/EverybodyDanceNow_reproduce_pytorch}}. 

\begin{table}[t]
\begin{subtable}[t] {0.48\linewidth}
\vspace{0pt}
\centering
\begin{tabular}{cc}
\toprule
Category & Number \\
\midrule
Ballet & 1165 \\
%\hline
Break & 3171 \\
%\hline
Cha & 4573 \\
%\hline
Flamenco & 2271 \\
%\hline
Foxtrot & 2981 \\
%\hline
Jive & 3765 \\
%\hline
Latin & 2205 \\
%\hline
Pasodoble & 2945 \\
%\hline
Quickstep & 2776 \\
%\hline
Rumba & 4459 \\
%\hline
Samba & 3143 \\
%\hline
Square & 5649 \\
%\hline
Swing & 3528 \\
%\hline
Tango & 3321 \\
%\hline
Tap & 2860 \\
%\hline
Waltz & 3046 \\
%\hline
\bottomrule
\end{tabular}  
\caption{Let's Dance Dataset.}\label{tbl:Dance}
\end{subtable}
\hfill
\begin{subtable}[t]{0.48\linewidth}  
\vspace{0pt}
\centering
\begin{tabular}{cc}
\toprule
Category & Number \\
\midrule
%\hline\hline
Clean Train& 1636 \\
%\hline
Clean Val& 146 \\
%\hline
Noisy Train& 656 \\
%\hline
Noisy Val& 74 \\
%\hline
\bottomrule
\end{tabular}
\caption{K-pop.}\label{tbl:K-pop}
\vspace{7pt}
\begin{tabular}{cc}
\toprule
Category & Number \\
\midrule
Pop & 4334 \\
%\hline
Rock & 4324 \\
%\hline
Instrumental & 4299 \\
%\hline
Electronic & 4333 \\
%\hline
Folk & 4346 \\
%\hline
International & 4341 \\
%\hline
Hip-Hop & 4303 \\
%\hline
Experimental & 4323 \\
\bottomrule
\end{tabular}
\caption{FMA.}\label{tbl:Music}
\end{subtable}
\caption{The detail of our datasets. All the datasets are cut into pieces of 5s. Number means the number of the pieces. For \textit{Let's Dance Dataset} and FMA, 70\% is for training, 5\% is for validation, and 25\% is for testing.}\label{tbl:main}
\end{table}

\section{Experiments}

% \begin{figure}[t]
% \centering
%   \includegraphics[scale=1]{./image/crop_CRNN.pdf}
%   \caption{CRNN uses a 2-layer RNN with GRU to summarise temporal patterns on the top of two dimensional 4-layer CNNs.}
% \label{fig:CRNN}
% \end{figure}

\subsection{Datasets}
\textbf{K-pop dataset.}
To build our dataset, we apply OpenPose~\cite{cao2018openpose,cao2017realtime,wei2016cpm} to some online videos to obtain the skeleton sequences. In total, We collected 60 videos about 3 minutes with a single women dancer and split these videos into two datasets. The first part with 20 videos is very noisy, as shown in Figure \ref{fig:K-pop}. This dataset is used to test the performance of the pose perceptual loss on noisy data. 18 videos of this part are for training, and 2 videos of this part are for evaluation. The second part with 40 videos is relatively clean and used to form our automatic dance video generation task. 37 videos of this part are for training, and 3 videos of this part are for evaluation.  % The skeleton sequences are scaled to a similar height: we first define one video's skeleton sequence as the standard sequence and used a weighted ratio to scale sequences of other videos to match the standard skeleton sequences. All the skeleton sequences are normalized to $[-1,1]$. 
The detail of this dataset is shown in Table \ref{tbl:K-pop}.

\textbf{Let's Dance Dataset.}
Castro et al.~\cite{castro2018let} released a dataset containing 16 classes of dance, presented in Table \ref{tbl:Dance}. The dataset provides information about human skeleton sequences for pose recognition. Though there are existing enormous motion datasets ~\cite{motion_dataset1,NTU,motion_dataset2} with skeleton sequences, we choose \textit{Let's Dance Dataset} to pre-train our ST-GCN for pose perceptual loss as dance is different with normal human motion. 
% For the preprocessing of the dataset, we split the multiple-person dance videos into single-person ones and normalize them to $[-1,1]$.

\begin{table}[t]
%\footnotesize
\centering
\begin{tabular}{lcc}
\toprule
\multicolumn{1}{c}{\bfseries Metric\mdseries}& Cross-modal& BRISQUE\\
\midrule  
Rand Frame & 0.151 & --\\
Rand Seq & 0.204 & --\\
$L_1$ & 0.312 & 40.66\\
Global D & 0.094 & 40.93\\
Local D & 0.068 & 41.46\\
Our model &\textbf{0.046}& \textbf{41.11}\\

\bottomrule
\end{tabular}
\caption{Results of our model and baselines. On the cross-modal evaluation, lower is better. For BRISQUE, higher is better. The details of the baselines are shown in Section \ref{subsection:Metrics}. }
\label{tbl:Metric}
\end{table}

\begin{figure*}[t]
\renewcommand{\arraystretch}{2}
\centering
    \begin{tabular}{c@{\hspace{0.4em}}c@{\hspace{0.4em}}c@{\hspace{0.4em}}c:c@{\hspace{0.4em}}c@{\hspace{0.4em}}c@{\hspace{0.4em}}c@{\hspace{0.4em}}}
         \frame{\includegraphics[width=0.23\columnwidth]{{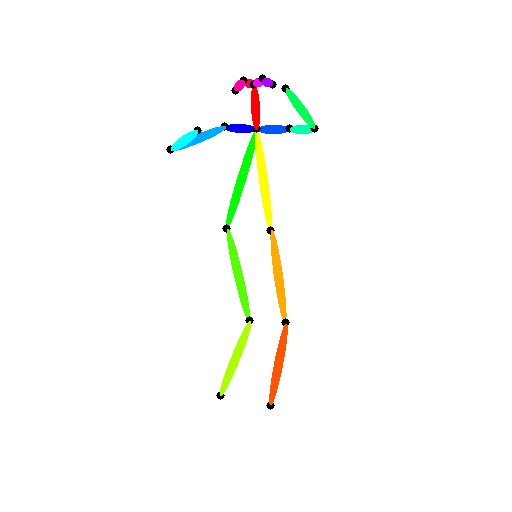}}} &
         \frame{\includegraphics[width=0.23\columnwidth]{{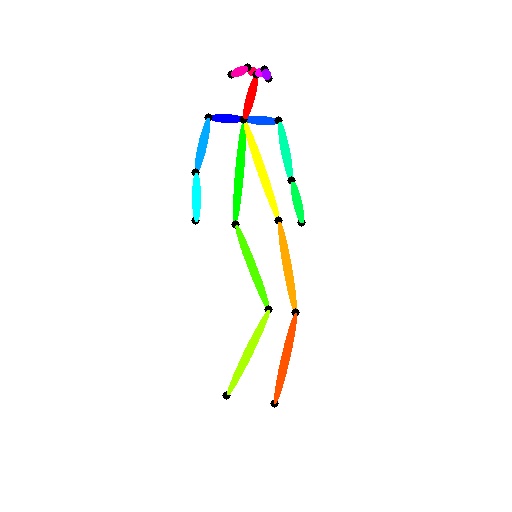}}} &
         \frame{\includegraphics[width=0.23\columnwidth]{{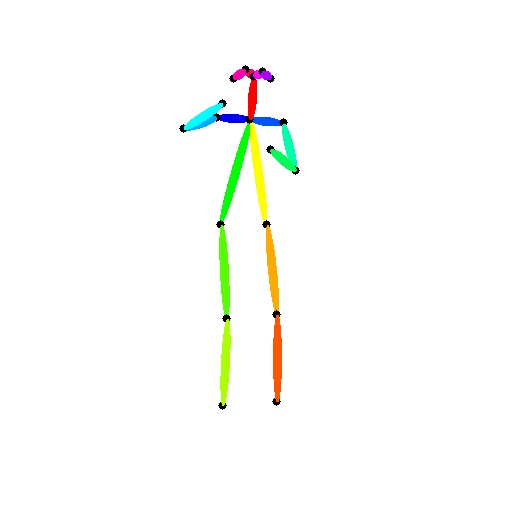}}} &
         \frame{\includegraphics[width=0.23\columnwidth]{{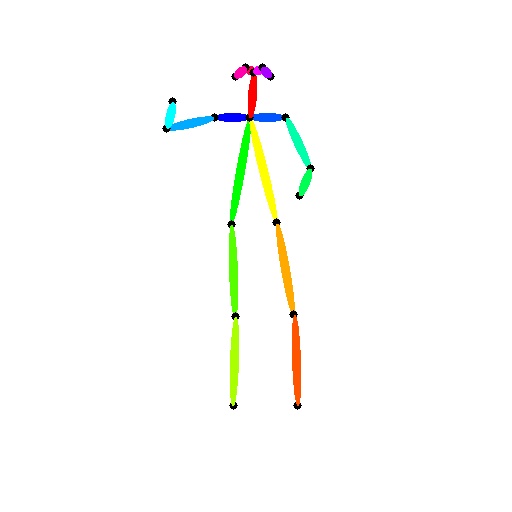}}} &
         \frame{\includegraphics[width=0.23\columnwidth]{{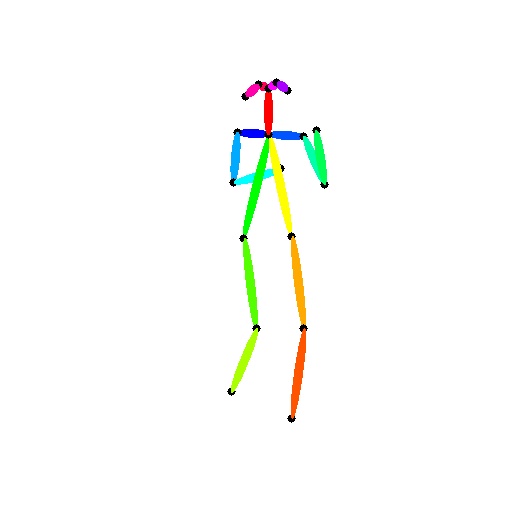}}} &
         \frame{\includegraphics[width=0.23\columnwidth]{{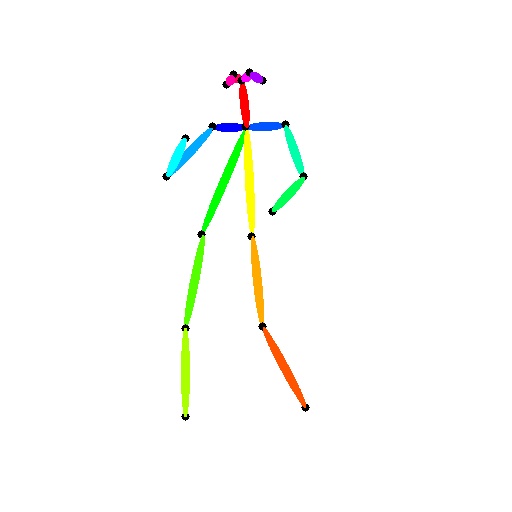}}} &
         \frame{\includegraphics[width=0.23\columnwidth]{{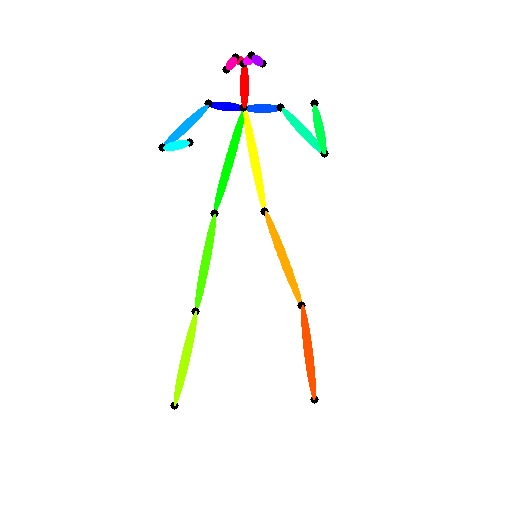}}} &
         \frame{\includegraphics[width=0.23\columnwidth]{{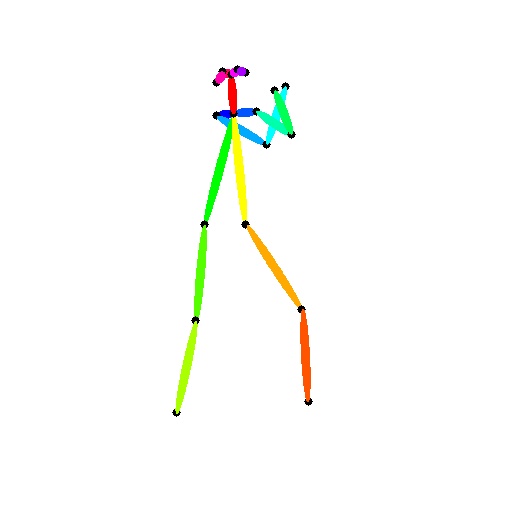}}} \\
         \includegraphics[width=0.23\columnwidth]{{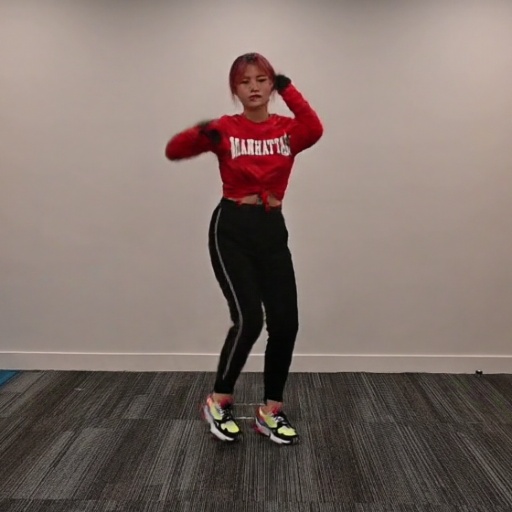}} &
         \includegraphics[width=0.23\columnwidth]{{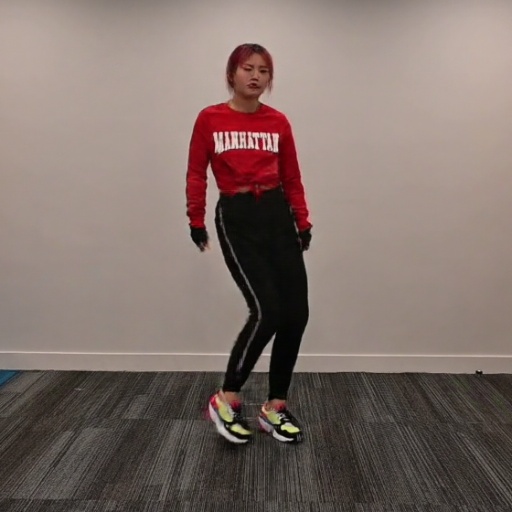}} &
         \includegraphics[width=0.23\columnwidth]{{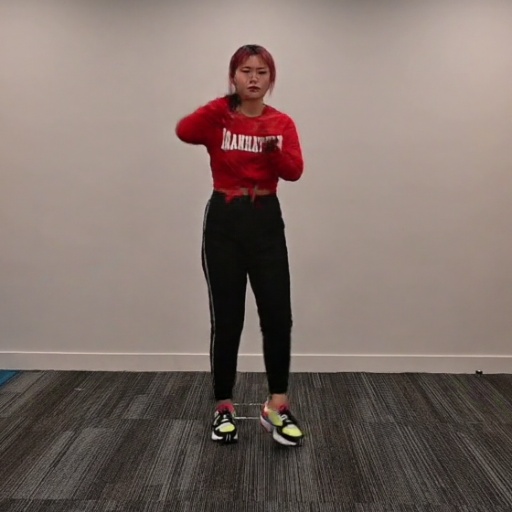}} &
         \includegraphics[width=0.23\columnwidth]{{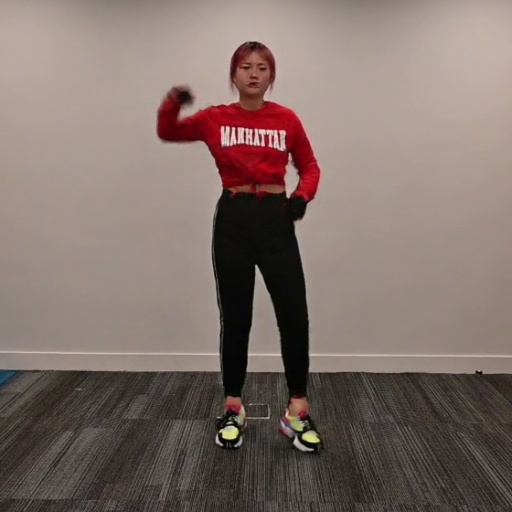}} &
         \includegraphics[width=0.23\columnwidth]{{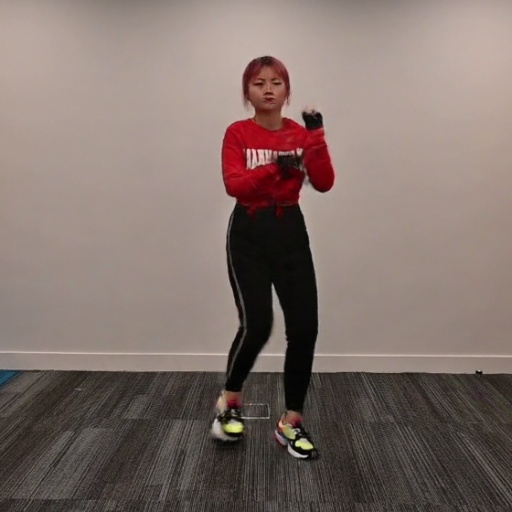}} &
         \includegraphics[width=0.23\columnwidth]{{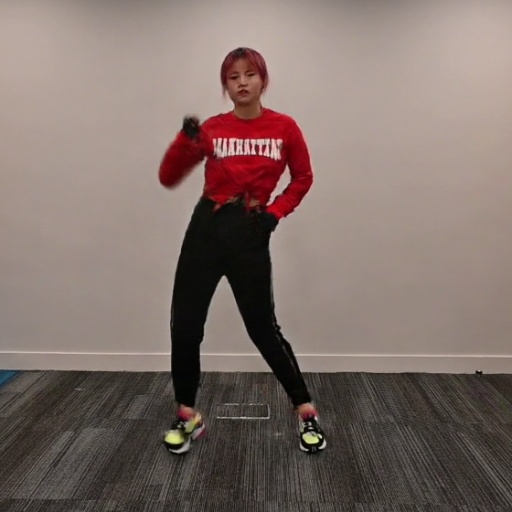}} &
         \includegraphics[width=0.23\columnwidth]{{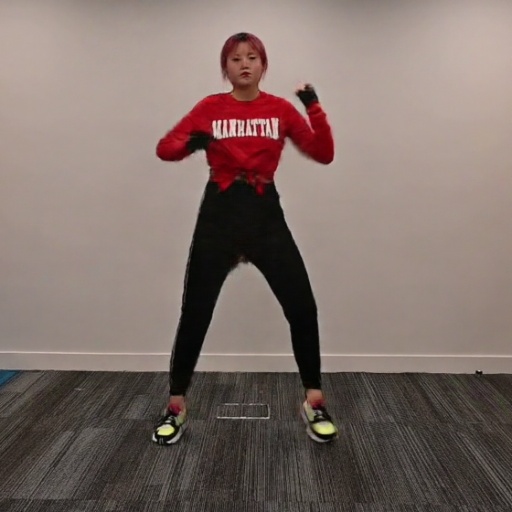}} &
         \includegraphics[width=0.23\columnwidth]{{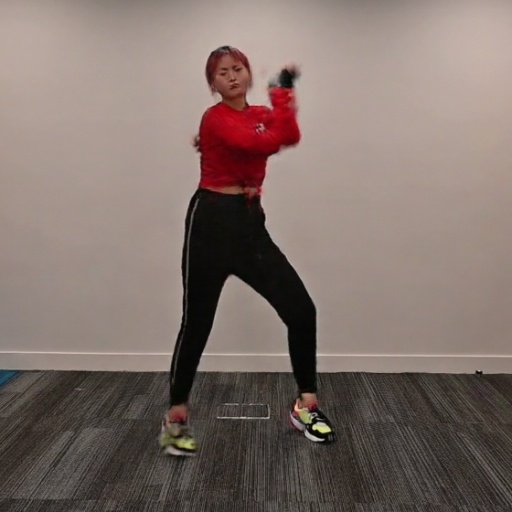}} \\
         \includegraphics[width=0.23\columnwidth]{{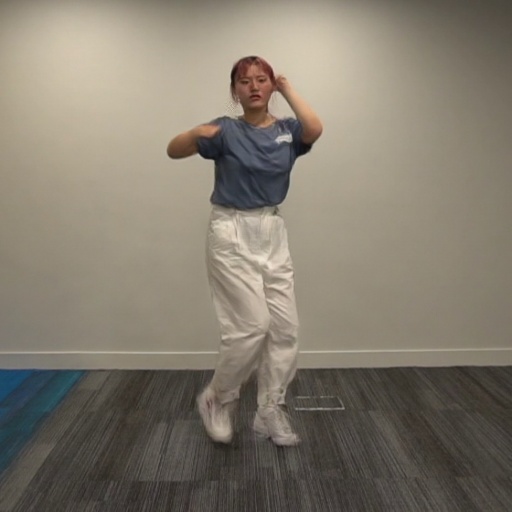}} &
         \includegraphics[width=0.23\columnwidth]{{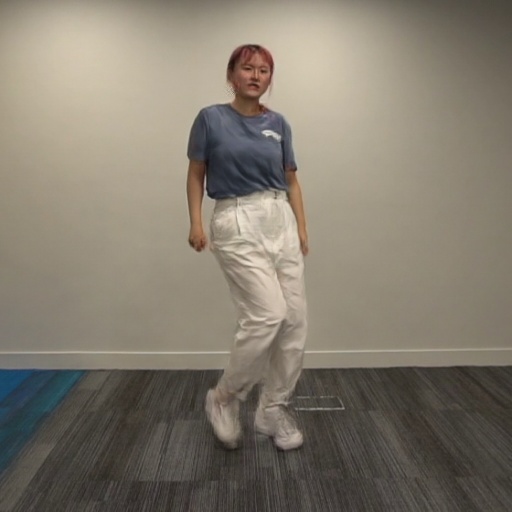}} &
         \includegraphics[width=0.23\columnwidth]{{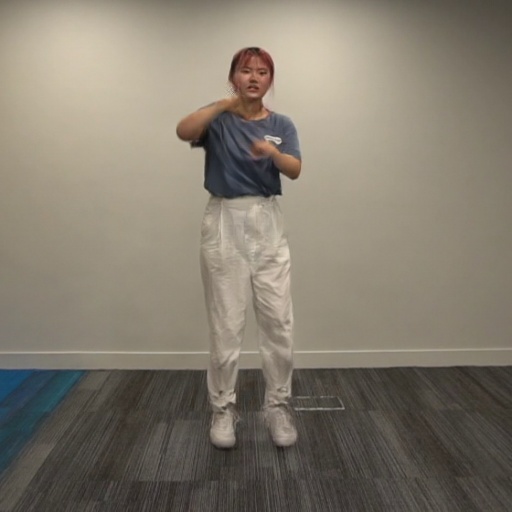}} &
         \includegraphics[width=0.23\columnwidth]{{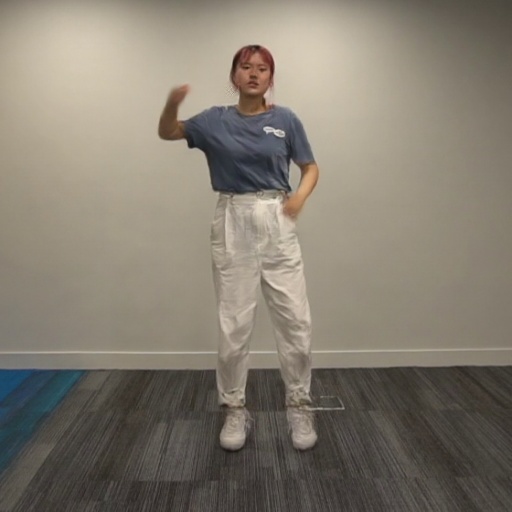}} &
         \includegraphics[width=0.23\columnwidth]{{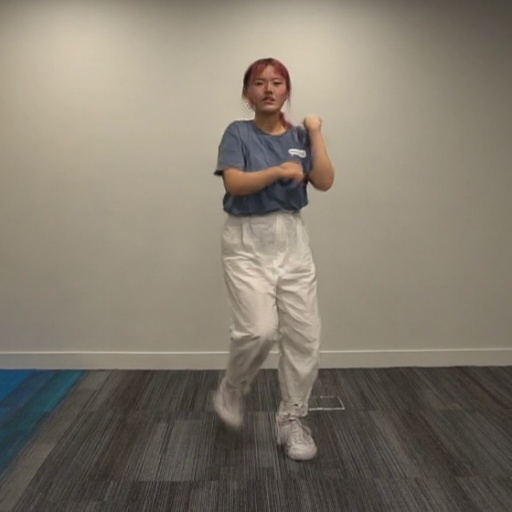}} &
         \includegraphics[width=0.23\columnwidth]{{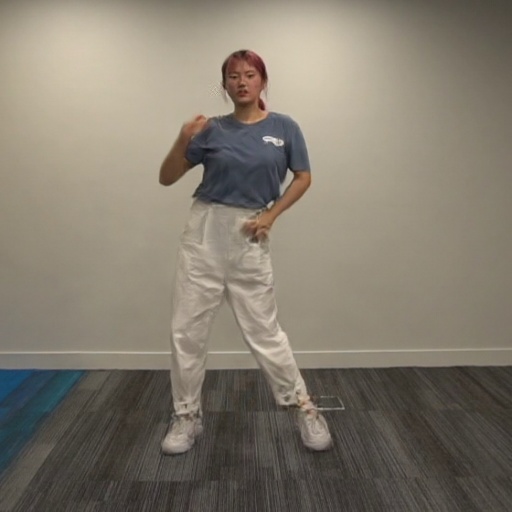}} &
         \includegraphics[width=0.23\columnwidth]{{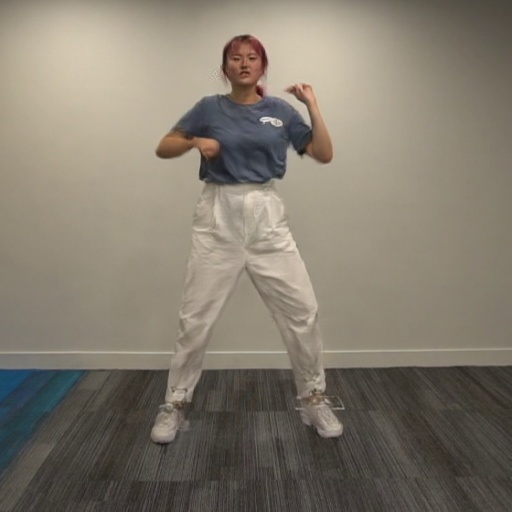}} &
         \includegraphics[width=0.23\columnwidth]{{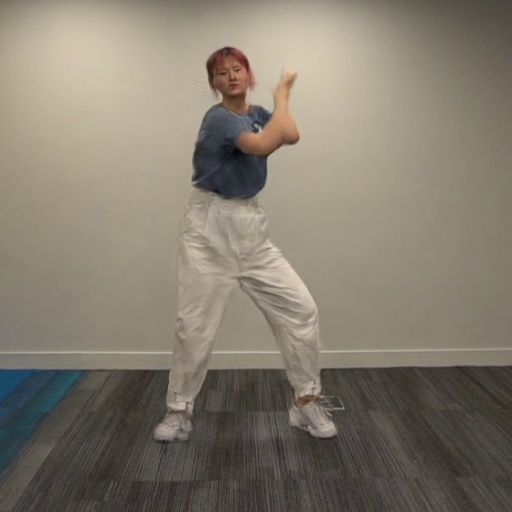}} \\
     \end{tabular} 
  \caption{Synthesized music video conditioned on the music \textsl{``LIKEY"} by \textsl{TWICE}. For each 5-second dance video, we show 4 frames. The top row shows the skeleton sequence, and the bottom row shows the synthesized video frames conditioned on different target videos.}
  %\label{fig:dn_Demo}
\end{figure*}

\textbf{FMA.}
For our cross-modal evaluation, the extraction of music features is needed. To achieve this goal, we adopt CRNN~\cite{ChoiFSC17} and choose the dataset Free Music Archive (FMA) to train CRNN. In FMA, genre information and the music content are provided for genre classification. The information of FMA is shown in Table \ref{tbl:Music}.

\subsection{Experimental Setup}
All the models are trained on an Nvidia GeForce GTX 1080 Ti GPU. For the first stage in our framework, the model is implemented in PyTorch~\cite{pytorch} and takes approximately one day to train for 400 epochs. For the hyperparameters, we set $V=18$, $T=50$, $t=5$, $K=16$, $S=16000$. For the self attention mechanism, we set $k=256,l=40$. For the loss function, the hyperparameters  $\{\lambda_l\}$ are set to be $[20,5,1,1,1,1,1,1,1]$ and $w_{GP}=1,w_P=1,w_{FM}=1,w_{L_1}=200.$ Though the weight of $L_1$ distance loss is relatively large, the absolute value of the $L_1$ loss is quite small. We used Adam~\cite{adam} for all the networks with a learning rate of 0.003 for the generator and 0.003 for the Local Temporal Discriminator and 0.005 for the Global Content Discriminator.

For the second stage that transfers pose to video, the model takes approximately three days to train, and the hyperparameters of it adopt the same as ~\cite{chan2018everybody}. For the pre-train process of ST-GCN and CRNN, we also used Adam~\cite{adam} for them with a learning rate of 0.002. ST-GCN achieves 46\% precision on \textit{Let's Dance Dataset}. CRNN is pretrained on the FMA, and the top-2 accuracy is 67.82\%.

\subsection{Evaluation}
\label{subsection:Metrics}
We will evaluate the following baselines and our model.
\begin{itemize}
\item \bm{$L_1$}. In this condition we just use $L_1$ distance to conduct the generator. 
\item \textbf{Global D}. Based on \bm{$L_1$}, we add a Global Content Discriminator. 
\item \textbf{Local D}.  Based on \textbf{Global D} , we add a Local Temporal Discriminator.
\item \textbf{Our model}. Based on \textbf{Local D}, we add pose perceptual loss. These conditions are used in Table \ref{tbl:Metric}.
\end{itemize}

\subsubsection{User Study}
\begin{figure}[t]
\centering
   \includegraphics[width=\columnwidth]{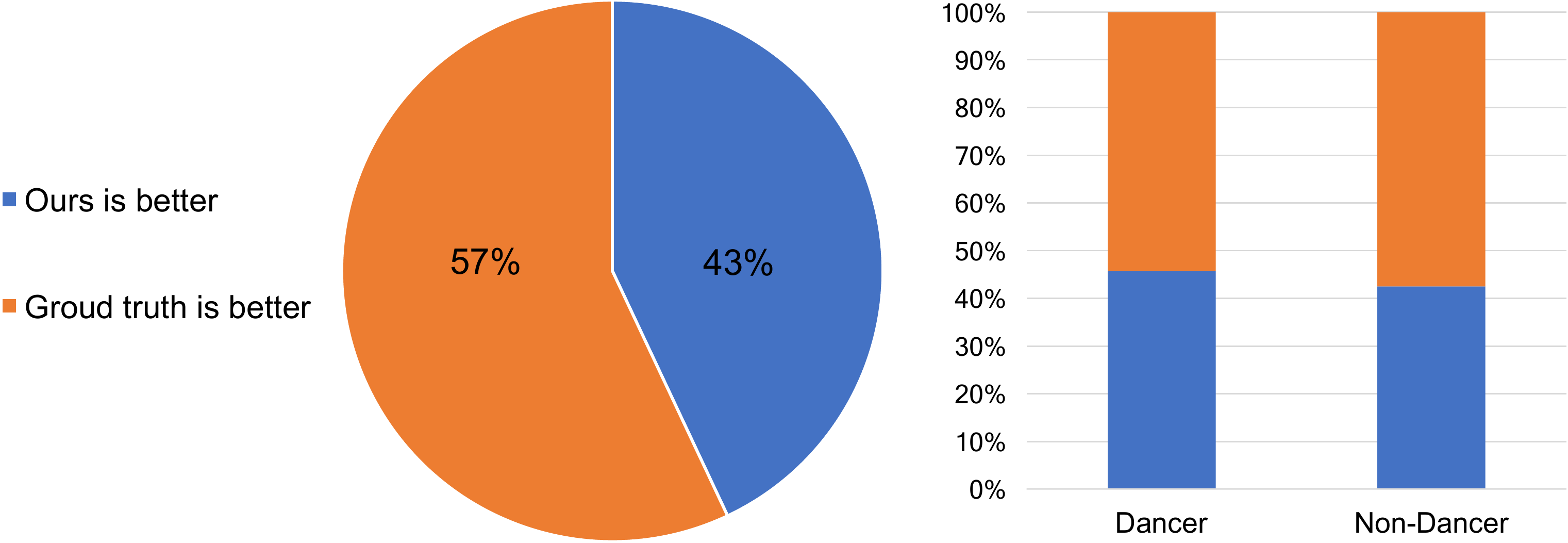}
   \caption{Results of user study on comparisons between the synthesized skeleton sequence and the ground truth. There are 27 participants in total, including seven dancers. In nearly half of the comparisons, users can not tell which skeleton sequence is better given the music. To make the results reliable, we make sure there is no unclean skeleton in the study.}
   %We apply a paired t-test to the results and gain $p=0.06> 0.05$, and thus, we can not determine that there is a significant difference in these preferences.
\label{fig:user}
\end{figure}

To evaluate the quality of the generated skeleton sequences (our main contributions), we conduct a user study comparing the synthesis skeleton sequence and the ground-truth skeleton sequence. We randomly sample 10 pairs sequences with different lengths and draw the sequences into videos. To make this study fair, we verify the ground truth skeletons and re-annotate the noisy ones. In the user study, each participant watches the video of the synthesis skeleton sequence and the video of the ground truth skeleton sequence in random order. Then the participant needs to choose one of the two options: 1) The first video is better. 2) The second video is better. As shown in Figure \ref{fig:user}, in 43.0\% of the comparisons, participants vote for our synthesized skeleton sequence. This user study shows that our model can choreograph at a similar level with real artists.

\begin{figure*}[t]
\begin{center}
   \includegraphics[width=\linewidth]{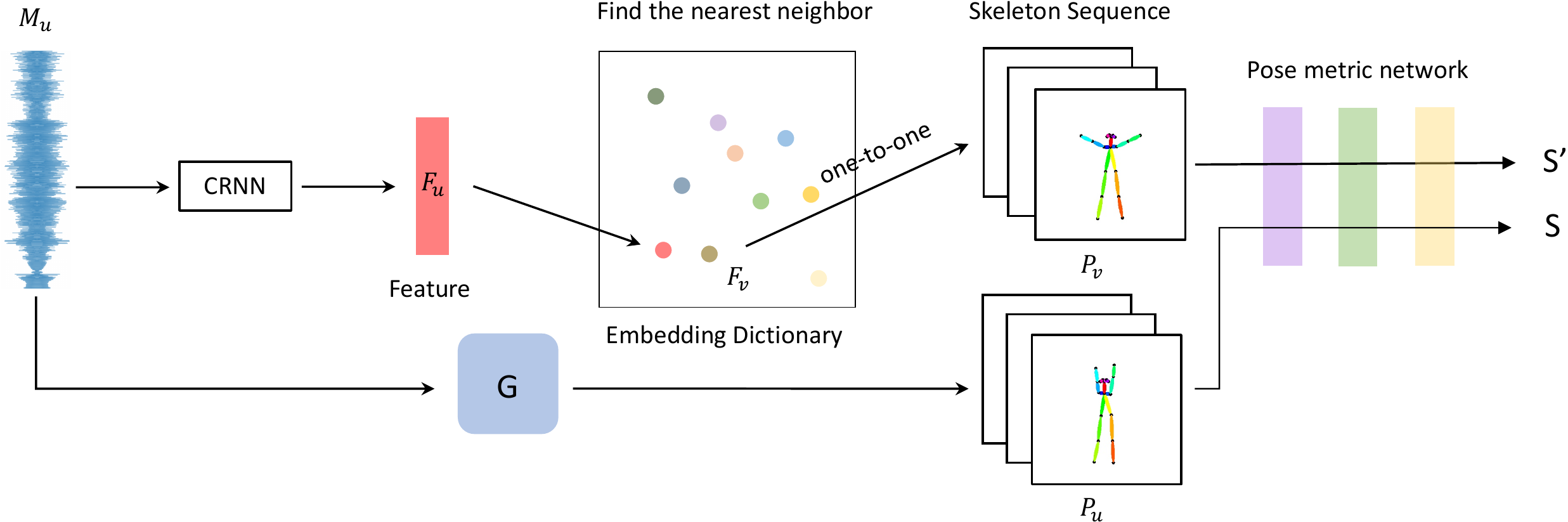}
\end{center}
   \caption{Cross-modal evaluation. We first project all the music pieces in the training set of the K-pop dataset into an embedding dictionary. We train the pose metric network based on the K-means clustering result of the embedding dictionary. For the K-means clustering, we choose K = 5, according to the Silhouette Coefficient. The similarity between $M_{u}$ and $P_{u}$ is measured by $\Arrowvert S - S' \Arrowvert^2$.}
\label{fig:Metric}
\vspace{1em}
\end{figure*}

\begin{figure*}[t]
\renewcommand{\arraystretch}{2}
\centering
    \begin{tabular}{c@{\hspace{0.4em}}c@{\hspace{0.4em}}c@{\hspace{0.4em}}c:c@{\hspace{0.4em}}c@{\hspace{0.4em}}c@{\hspace{0.4em}}c@{\hspace{0.4em}}}
         \frame{\includegraphics[width=0.23\columnwidth]{{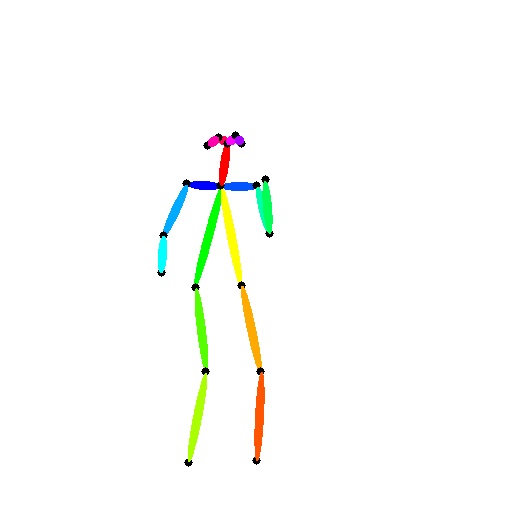}}} &
         \frame{\includegraphics[width=0.23\columnwidth]{{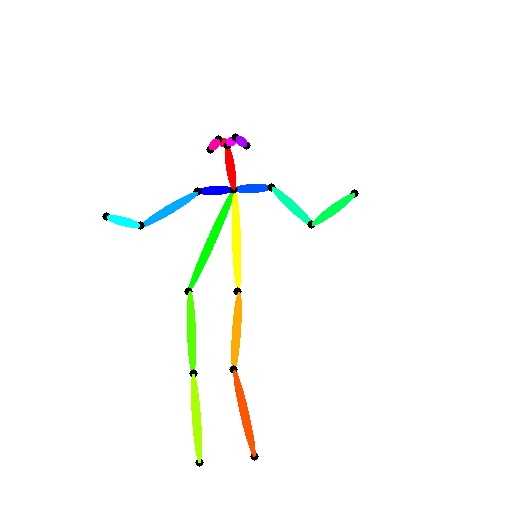}}} &
         \frame{\includegraphics[width=0.23\columnwidth]{{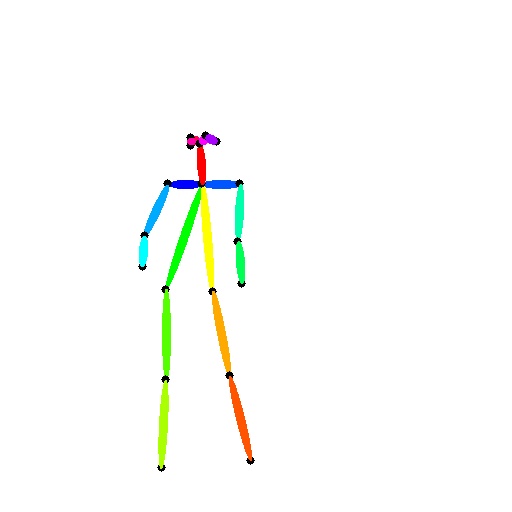}}} &
         \frame{\includegraphics[width=0.23\columnwidth]{{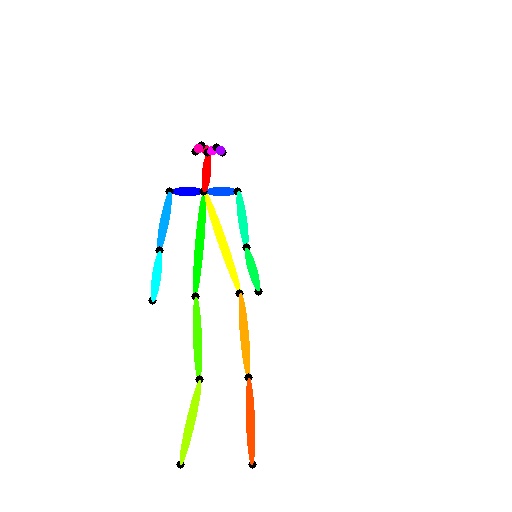}}} &
         \frame{\includegraphics[width=0.23\columnwidth]{{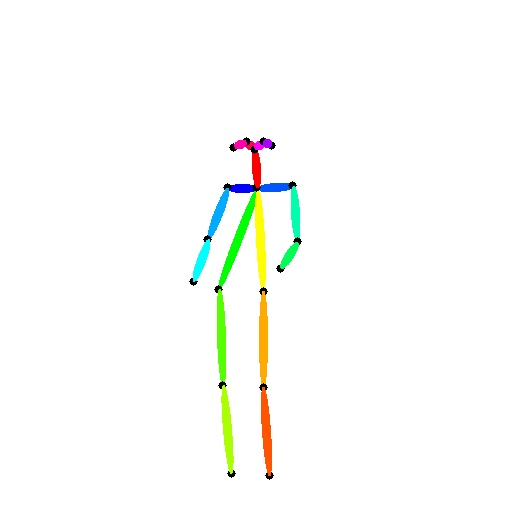}}} &
         \frame{\includegraphics[width=0.23\columnwidth]{{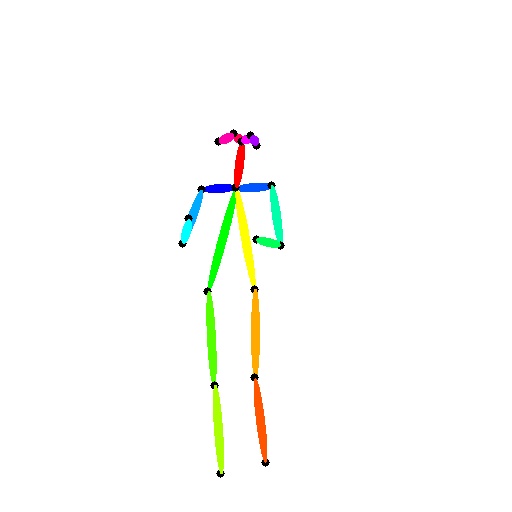}}} &
         \frame{\includegraphics[width=0.23\columnwidth]{{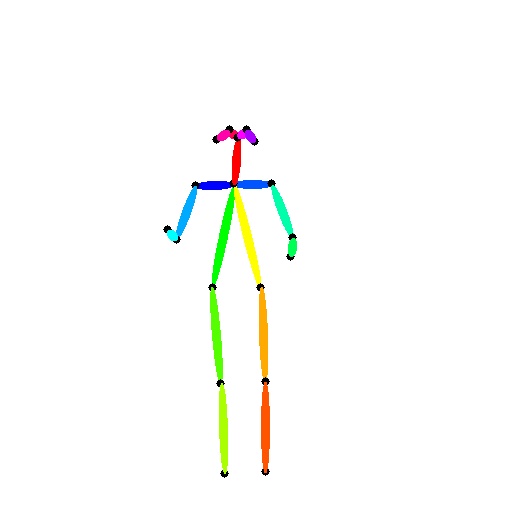}}} &
         \frame{\includegraphics[width=0.23\columnwidth]{{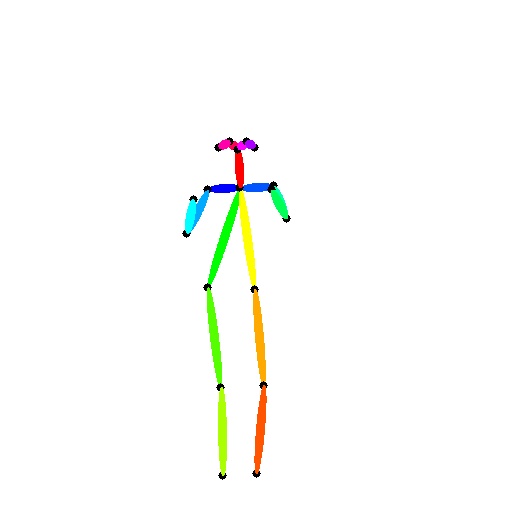}}}\\
         \includegraphics[width=0.23\columnwidth]{{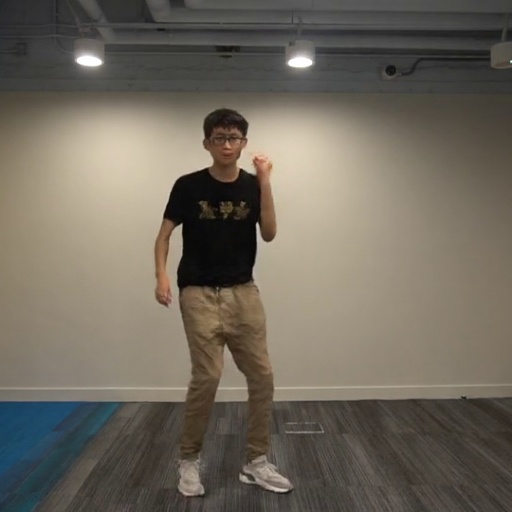}} &
         \includegraphics[width=0.23\columnwidth]{{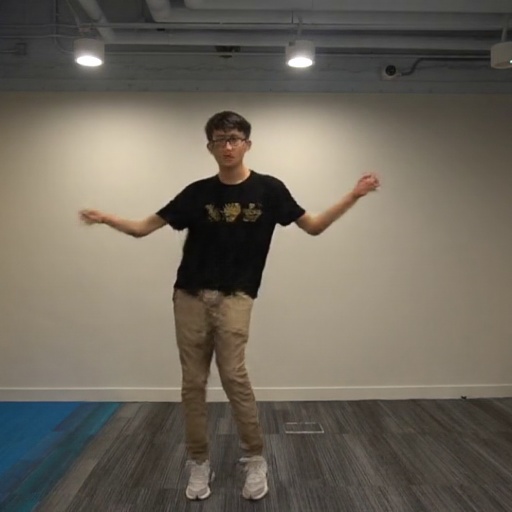}} &
         \includegraphics[width=0.23\columnwidth]{{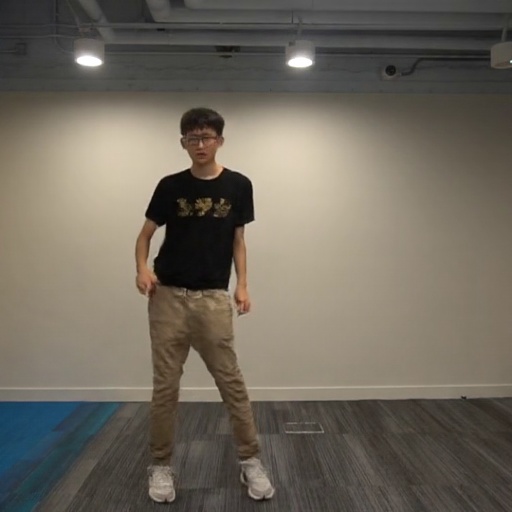}} &
         \includegraphics[width=0.23\columnwidth]{{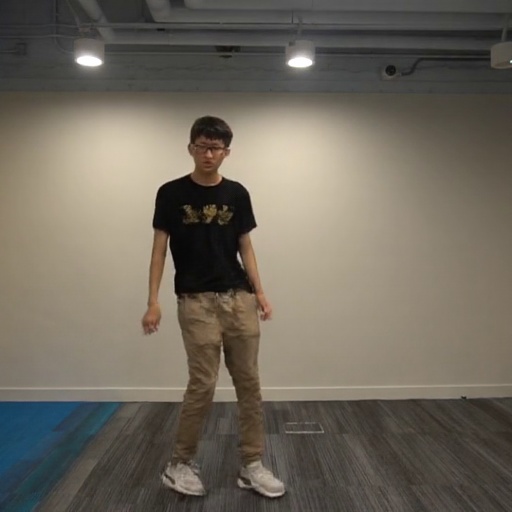}} &
         \includegraphics[width=0.23\columnwidth]{{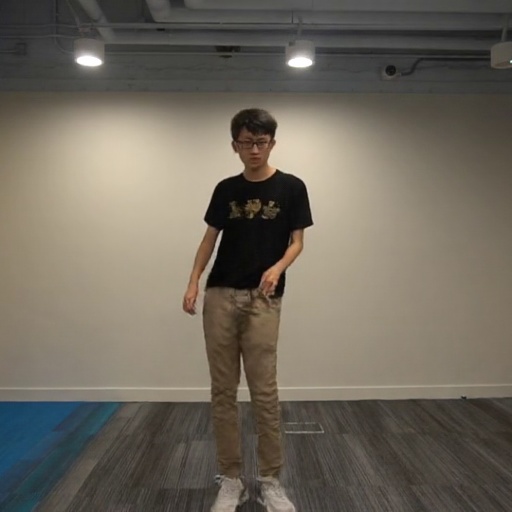}} &
         \includegraphics[width=0.23\columnwidth]{{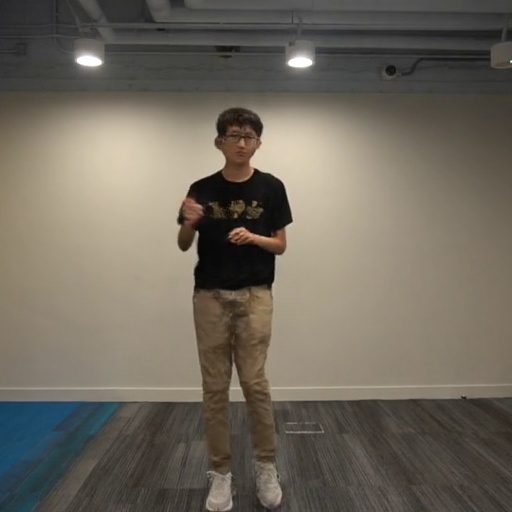}} &
         \includegraphics[width=0.23\columnwidth]{{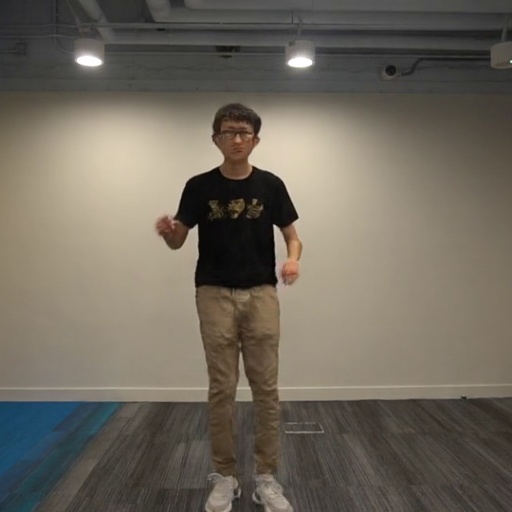}} &
         \includegraphics[width=0.23\columnwidth]{{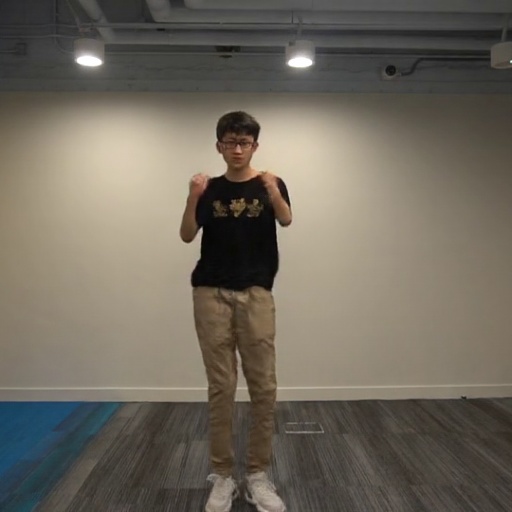}} \\
     \end{tabular} 
  \caption{Our synthesized music video with a male student as a dancer.}
\end{figure*}

\subsubsection{Cross-modal Evaluation}
It is challenging to evaluate if a dance sequence is suitable for a piece of music. To our best knowledge, there is no existing method to evaluate the mapping between music and dance. Therefore, we propose a two-step cross-modal metric, as shown in Figure \ref{fig:Metric}, to estimate the similarity between music and dance. 

Given a training set $X=\{(P, M)\}$ where $P$ is a dance skeleton sequence and $M$ is the corresponding music. Then with a pre-trained music feature extractor $E_{m}$~\cite{ChoiFSC17}, we aggregate all the music embeddings $F=\{E_m(M),M\in X\}$ in an embedding dictionary.

The input to our evaluation is music $M_u$. With our generator $G$, we can get the synthesized skeleton sequence $P_{u}=G(M_u)$. The first step is to find a skeleton sequence that represents the music $M_{u}$. We first obtain the music feature $F_{u}$ by $F_{u}=E_{m}(M_{u})$. Then let $F_{v}$ be the nearest neighbor of $F_{u}$ in the embedding dictionary. In the end, we use its corresponding skeleton sequence $P_{v}$ to represent the music $M_{u}$. The second step is to measure the similarity between two skeleton sequences with the novel metric learning objective based on a triplet architecture and Maximum Mean Discrepancy, proposed by Coskun et al.~\cite{CoskunTCNT18}. More implementation details about this metric will be shown in supplement materials.

%CQF%The training procedure of this pose metric network needs the label of category, while the only label for the skeleton sequences is the corresponding pieces of music whose genre is mainly K-pop. To label the dance sequence, we apply K-means clustering to the embedding space. Then we have K clusters in which the music is similar, and we use the categories of clusters to label the skeleton sequences. More details about this metric will be shown in supplement materials.

\subsubsection{Quantitative Evaluation}
To evaluate the quality of results of the final method in comparison to other conditions, Chan et al.~\cite{chan2018everybody} propose to make a transfer between the same video since there is no reference for the synthesized frame and use SSIM~\cite{SSIM} and LPIPS~\cite{LPIPS} to measure the videos. For our task, such metrics are useless because there are no reference frames for the generated dance video. So we apply BRISQUE~\cite{BRISQUE}, which is a no-reference Image Quality Assessment to measure the quality of our final generated dance video. 

As shown in Table \ref{tbl:Metric}, by utilizing the Global Content Discriminator and the Local Temporal Discriminator, even for a single frame result, the score is better. For the addition of the pose perceptual loss, the poses become plausible, and then transferring the diverse poses to the frames may lead to the decline of the score. Furthermore, more significant differences can be observed in our video. To validate our proposed evaluation, we also try two random conditions: 
\begin{itemize}
\item \textbf{Rand Frame}. Randomly select 50 frames from the training dataset for the input music instead of feeding the music into the generator. 
\item \textbf{Rand Seq}. Randomly select a skeleton sequence from the training dataset for the input music instead of feeding the music into the generator. 

To make the random results stable, we make ten random processes and get the average score.
\end{itemize}

\section{Conclusion}
We have presented a two-stage framework to generate dance videos, given any music. With our proposed pose perceptual loss, our model can be trained on dance videos with noisy pose skeleton sequence (no human labels). Our approach can create arbitrarily long, good-quality videos. We hope that this pipeline of synthesizing skeleton sequence and dance video combining with pose perceptual loss can support more future work, including more creative video synthesis for artists.

% However, it suffers from several limitations.
% Limited to the number of dance videos that we collect, we mainly work on K-pop with a single person, which is just a kind of dance. Further work could focus on improving results by obtaining much more data and generate arbitrary sort of dancing with the arbitrary number of people.

% Though our model can generate diverse dance movements according to different music, the synthesized dance is still not as good as real dance. Specifically, for every K-pop music, there are some unique dance movements, and our model can not capture all these unique movements. Future work could focus on making further improvement on the creativity of the dance generator.

% If such problems are solved, which we believe to be possible, this task has a lot of potential applications.

{\small
\bibliographystyle{ieee_fullname}
\bibliography{main}
}

\end{document}